\documentclass[10pt,twocolumn,letterpaper]{article}

\usepackage{iccv}
\usepackage{times}
\usepackage{epsfig}
\usepackage[caption=false]{subfig}
\usepackage{graphicx}
\usepackage{amsmath}
\usepackage{amssymb}
\usepackage{adjustbox}
\usepackage{comment}

\usepackage[breaklinks=true,bookmarks=false]{hyperref}

\usepackage{cleveref}
\usepackage{pifont}
\crefname{figure}{Fig.}{Figs.}
\crefname{table}{Table}{Tables}
\crefname{section}{Section}{Sections}
\newcommand{\xmark}{\ding{55}}

\iccvfinalcopy 

\def\iccvPaperID{****} 

\ificcvfinal\fi
\def\TransformerName{Gated Attention\;}
\begin{document}

\title{A Gated Attention Transformer for Multi-Person Pose Tracking}
\author{
     Andreas Doering$^{1,2}$  \qquad
     Juergen Gall$^{1,2}$ \\
     {\normalsize
     $^1$University of Bonn} \qquad
     {\normalsize$^2$Lamarr Institute for Machine Learning and Artificial Intelligence}
}

\maketitle
\ificcvfinal\thispagestyle{empty}\fi

\begin{abstract}
Multi-person pose tracking is an important element for many applications and requires to estimate the human poses of all persons in a video and to track them over time. The association of poses across frames remains an open research problem, in particular for online tracking methods, due to motion blur, crowded scenes and occlusions. To tackle the association challenge, we propose a Gated Attention Transformer. The core aspect of our model is the gating mechanism that automatically adapts the impact of appearance embeddings and embeddings based on temporal pose similarity in the attention layers. In order to re-identify persons that have been occluded, we incorporate a pose-conditioned re-identification network that provides initial embeddings and allows to match persons even if the number of visible joints differ between frames. We further propose a matching layer based on gated attention for pose-to-track association and duplicate removal. We evaluate our approach on PoseTrack 2018 and PoseTrack21.

\end{abstract}

\section{Introduction}
\begin{figure*}[t]
\begin{center}
\includegraphics[width=\linewidth]{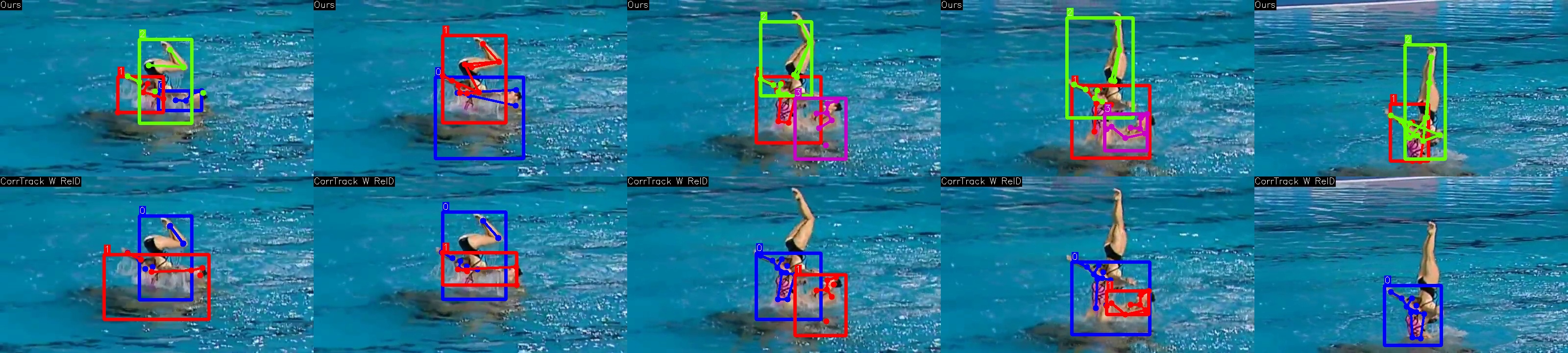}
\end{center}
   \caption{Qualitative examples of our proposed method on the PoseTrack21 dataset. The first row contains visual tracking results of our method and the second row shows visualizations of CorrTrack with ReID \cite{posetrack21}. }
\label{fig:qualitative}
\vspace{-5mm}
\end{figure*}

Multi-person pose tracking is highly relevant for a wide range of applications such as virtual reality, autonomous driving or sports analysis and requires to accurately estimate and track the human poses of all persons throughout a video. Despite of the recent progress in multi-person pose tracking~\cite{STAF,doering18,mipal,learning_to_track_instances,15_keypoints,simple_baseline,rafi2020selfsupervised,learning_dynamics_via_graph,combining_detection_and_tracking}, the task remains very challenging due to camera motion, motion blur, occlusions, and a high variety in pose and scale~\cite{posetrack21}. Consequently, a tracking approach must be robust to detection errors and ambiguities. In particular, the assignment of highly occluded persons in unusual poses is very difficult as shown in Fig.~\ref{fig:qualitative}, where three persons perform a gymnastic exercise in water. For instance, the green bounding box intersects with three persons and the respective pose overlaps with the keypoints of the other persons. This poses a challenge, especially for on-line methods as assignments will be made once a new frame arrives.

Related works such as \cite{15_keypoints,rafi2020selfsupervised,learning_dynamics_via_graph,simple_baseline,zhou2020temporal} try to tackle these challenges by generating future poses from a track's history, which are then matched with detections based on pose similarities, \eg, based on Object Keypoint Similarity (OKS) \cite{rafi2020selfsupervised,simple_baseline,zhou2020temporal} or a pose-based matching layer \cite{15_keypoints}. Other works such as \cite{combining_detection_and_tracking} process each sequence in an off-line fashion, which is not feasible for real-time applications. As these works mainly rely on pose-based similarities for matching, these methods tend to fail to re-identify tracks that have been occluded for longer periods of time or undergo high pose deformations. 

In our work, we thus focus on learning the association of detected persons to tracks in an on-line fashion and propose an approach that leverages the estimated poses, bounding boxes, and the appearance of the detected persons to assign them to previous tracks or initialize new tracks.  
Since we can neither rely solely on appearance-based features nor pose similarities due to multiple instances with similar appearance, changing camera views or scene switches, which often occur in in-the-wild sequences, we introduce two types of embeddings. The detection and track embeddings are based on appearance and used to measure the appearance similarity between detections and previous tracks. The additional edge embeddings directly encode the pose similarity between a pose and a track based on estimated poses and bounding boxes. While the pose similarity is a strong prior for tracking, it fails in case of fast motion or if the person disappeared for some frames due to occlusion or being outside of the camera view. We thus propose a gated attention transformer that combines and weights the attention matrices of both embedding types.    
All three embeddings are updated by a gated attention decoder and a final matching layer assigns the detections to the tracks. The matching layer also removes duplicates, \ie, multiple detections for the same persons, and initiates new tracks. Furthermore, we employ a pose-conditioned re-identification model where the appearance features are normalized based on the heatmaps of the detected keypoints.

We evaluate our approach on the challenging PoseTrack 2018 \cite{posetrack18} and PoseTrack21~\cite{posetrack21} datasets where it achieves state-of-the-art results. 
In summary, we propose i) a gated attention transformer that combines pose and appearance similarity in a novel way for pose-to-track association and ii) a novel matching layer for pose-to-track association and duplicate removal.   

\section{Related Work}
We briefly discuss some related works for person re-identification and multi-person pose tracking.

\noindent\textit{Person Re-Identification:}
Methods for person re-identification aim to generate robust feature representations of a query person, which allows to re-identify each instance of the particular person. These methods can be divided into different categories such as re-identification based on global features \cite{bagoftricks,HermansBeyer2017Arxiv, chen2023beyond, Wieczorek2021OnTU}, part-based approaches \cite{pcb,DIP,sun2019}, prior-based, \ie, pose- or mask-guided \cite{gao2020pose,sarfraz18,8693885,Wei2017GLADGD,DBLP:journals/corr/abs-1804-03864} and video-based \cite{Chen2018VideoPR, Chen2021KeypointMP, li2019} re-identification. 
Part-based models \cite{pcb,DIP,sun2019} divide each image into several parts and extract distinct part-based features. Features for each part are either trained individually \cite{pcb} or further re-combined to obtain visibility-aware features \cite{sun2019}. Li \etal \cite{DIP} propose to learn discriminative implicit parts (DiP) based on a vision transformer (ViT) \cite{vit} architecture, which tokenizes each image into equally sized patches. 
Pose- \cite{gao2020pose,sarfraz18,8693885,Wei2017GLADGD} or mask-guided \cite{DBLP:journals/corr/abs-1804-03864} approaches aim to suppress background-noise by learning occlusion- and instance-aware features that are more robust in crowded scenarios with occlusions. In this work, we propose Spatially Adaptive Pose DEnormalization (SPAPDE) layers for pose-guided re-identification and use it in the proposed gated attention transformer for multi-person pose tracking.

\noindent\textit{Multi-Person Pose Tracking:}
Existing works on multi-person pose tracking can be divided into two categories: top-down methods \cite{learning_dynamics_via_graph,combining_detection_and_tracking,15_keypoints,rafi2020selfsupervised,posetrack21,zhou2020temporal,10.1007/978-3-030-11012-3_19} and bottom-up methods \cite{STAF,mipal,learning_to_track_instances,doering18}. Former approaches employ a person detector and estimate the pose for every detected person individually based on temporal-context. Most methods employ a pose-warping \cite{rafi2020selfsupervised,posetrack21,10.1007/978-3-030-11012-3_19} scheme that warps tracked poses into the next frame or directly predict poses based on the tracklet history \cite{learning_dynamics_via_graph,15_keypoints,zhou2020temporal,jin2019,doering18}, which are then matched with detected poses using greedy or Hungarian matching.
In \cite{combining_detection_and_tracking}, an offline approach has been proposed that merges multiple overlapping fixed-lengths tracklets into tracks based on bipartite matching and Dijkstra’s algorithm \cite{Dijkstra1959}. Bottom-up approaches \cite{STAF,mipal,jin2019,doering18}, on the other hand, predict all keypoints within an image simultaneously and generate tracks by solving spatio-temporal graphs between detected keypoints. For instance, \cite{STAF,doering18,mipal} generate spatio-temporal vector fields,
while various spatio-temporal embeddings for the association of keypoints and tracks are proposed in \cite{jin2019}. In contrast, person instances are tracked in \cite{learning_to_track_instances} using a semi-supervised approach based on video instance correspondences. 
In this work, we propose a gating mechanism that automatically adapts the impact of appearance embeddings and edge embeddings, which are a strong prior and encode only pose and bounding box similarity, in the attention layers.

\begin{figure*}[t]
\begin{center}
   \includegraphics[width=0.65\textwidth]{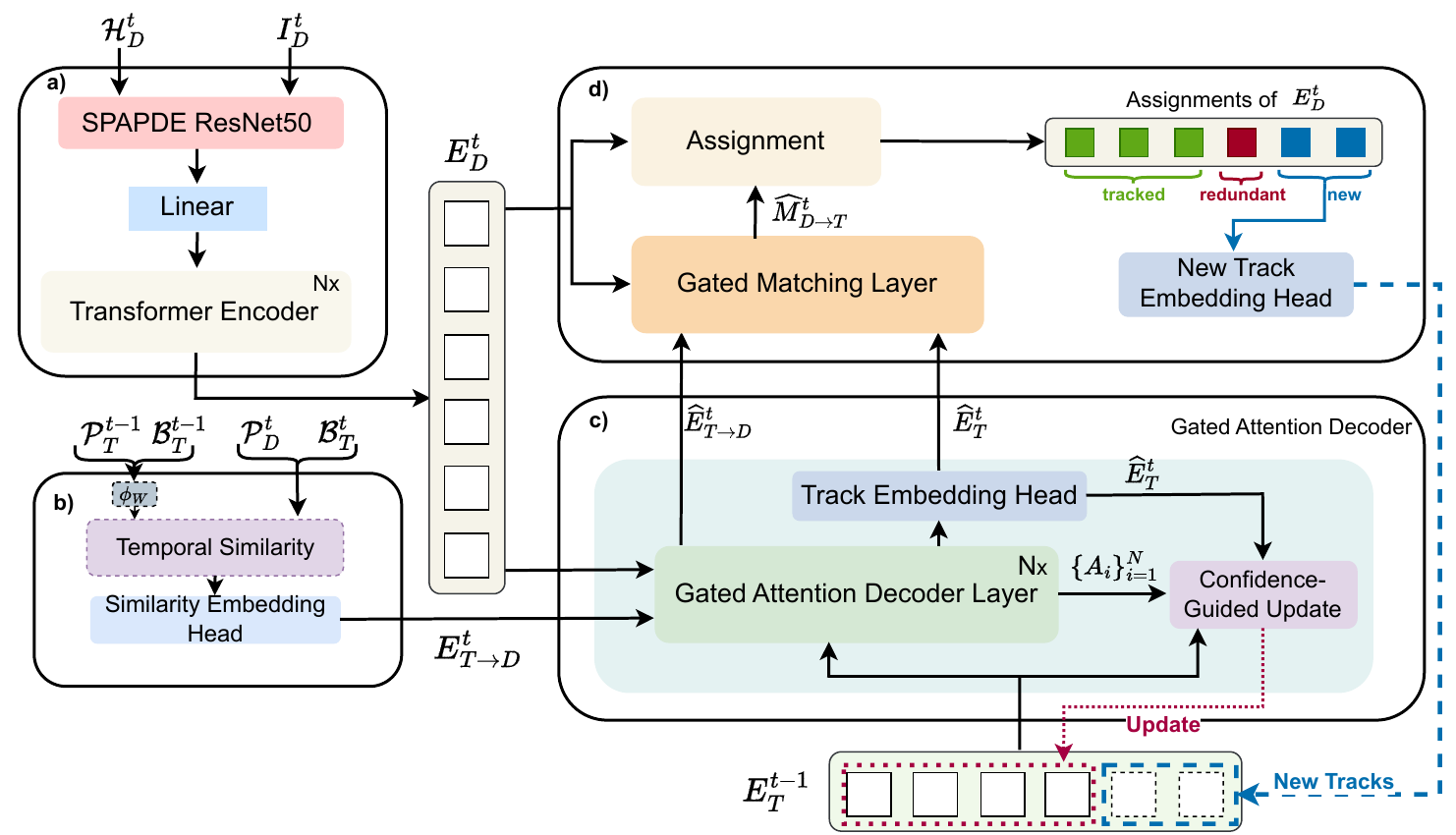}
\end{center}
\caption{The proposed multi-person pose tracking architecture entails the following steps:
a) Given a set of person crops and their respective keypoint heatmaps at time frame $t$, we compute pose-conditioned appearance features and feed them into N transformer encoder stages. This gives an embedding for each detection $E^t_D$, which will be used to measure the similarity to previous tracks. 
b) We also compute spatial similarities between tracks and detected persons by means of Intersection over Union (IoU) and Object Keypoint Similarity (OKS), which we then encode into a pose similarity embedding $E^t_{T \rightarrow D}$. This serves as a strong prior for matching. 
c) The gated attention decoder takes the embeddings of the previous tracks $E^{t-1}_T$, the detection embeddings $E^t_D$ and the pose similarity embeddings $E^t_{T \rightarrow D}$ as input and updates the embeddings. It adaptively weights the spatial prior $E^t_{T \rightarrow D}$ and the appearance similarity between $E^{t-1}_T$ and $E^t_D$.      
d) Finally, the matching stage assigns detections to tracks, removes redundant detections and initializes new tracks. }
\label{fig:architecture}
\end{figure*}
\section{\TransformerName Transformer for Multi-Person Pose Tracking}

On-line methods for multi-person pose tracking often follow the tracking-by-detection paradigm \cite{15_keypoints, detect_n_track, doering18, STAF, mipal, simple_baseline, offset_guided_networks, rafi2020selfsupervised} and usually suffer from ambiguities and occlusions. An example is shown in the second row of \cref{fig:qualitative} where the blue id jumps between two persons. In order to make the matching between detections and previously tracked persons more robust, we propose a gated attention transformer that directly learns the matching by a gated matching layer. It combines appearance features and encoded temporal person similarities. Since the importance of appearance and pose similarity varies within a video and between videos, in particular when a person has been occluded for a few frames, the gated attention decoder and the matching layer use a gating mechanism to update the embeddings of the new poses and previous tracks and to match poses to tracks. For example, if there are several very similar looking persons in a frame, the pose similarity can guide the update of the appearance embedding. Vice versa, the update will be driven by the appearance embeddings if there is no spatial proximity between tracks and detections.    
Our approach for multi-person pose tracking is illustrated in \cref{fig:architecture}.   

We assume that the human poses are extracted for a new frame $t$ by a standard multi-person pose estimator where we utilize the detector from \cite{posetrack21} for a fair comparison. Specifically, for a given frame at time $t$, our network takes as input the set of estimated bounding boxes $\mathcal{B}_D^t$ and poses $\mathcal{P}_D^t$, and additionally detected keypoint heatmaps $\mathcal{H}^t_D$ extracted from the image crops $\mathcal{I}_D^t$ of detected persons.  
From the heatmaps and image crops, an embedding $E^t_D$ for re-identification is computed as described in \cref{sec:encoder} (\cref{fig:architecture}a). To measure pose and spatial proximity between detections and tracks, similarities to the bounding boxes $\mathcal{B}_T^{t-1}$ and poses $\mathcal{P}_T^{t-1}$ of the last frame of each track are computed, which results in the edge embedding $E_{T\rightarrow D}^t$ (\cref{fig:architecture}b) between a track $T$ and a detection $D$, which will be described in \cref{sec:encoder} as well. We denote $E_{T\rightarrow D}^t$ as pose similarity embedding. 
Given both embeddings $E^t_D$ and $E_{T\rightarrow D}^t$ as well as the track embeddings $E_T^{t-1}$ that have been estimated in the previous frame $t{-}1$, the proposed \emph{Gated Attention Decoder} (\cref{fig:architecture}c) and the \emph{Gated Matching Layer} (\cref{fig:architecture}d) assign the current detections to previous tracks, update the track embeddings, and initialize new tracks. Both will be described in the following \cref{sec:twostream}.

\subsection{Gated Attention Decoder}\label{sec:twostream}
\begin{figure}[t]
\begin{center}
   \includegraphics[trim=30 0 30 0, clip,width=1\linewidth]{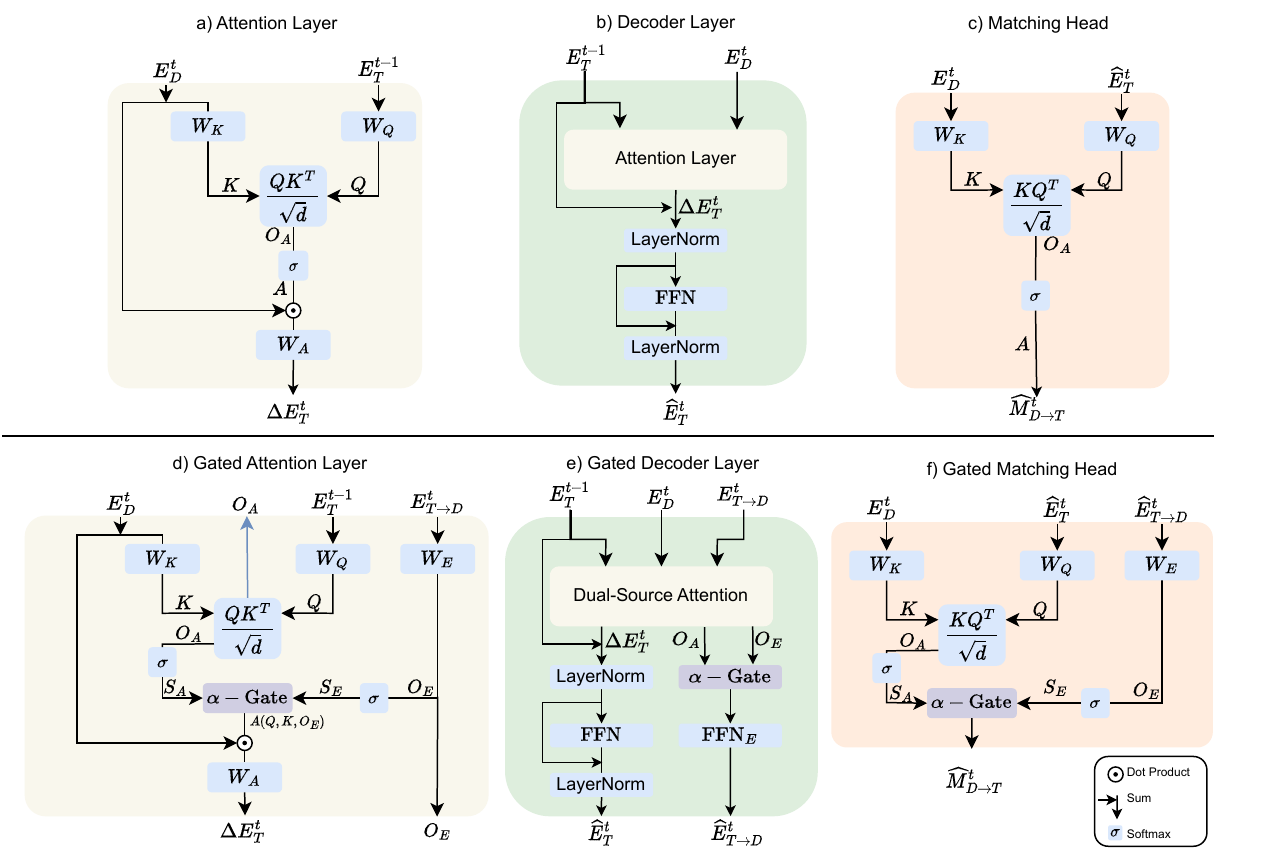}
\end{center}
   \caption{Illustration of our proposed d) Gated Attention layer, e) Gated Decoder layer and f) Gated Matching layer. The top row shows a vanilla implementation without gating for each layer: a) Standard Attention layer with $E_D^t$ as keys and $E_T^{t-1}$ as queries. b) The Decoder layer takes as input the detection embeddings $E_D^t$ and track embeddings $E_T^{t-1}$ and updates the track embeddings $\hat{E}_T^{t}$ using the Attention layer. c) The Matching layer predicts an assignment matrix $\widehat{M}_{D\rightarrow T}^t$ of detections to tracks. The proposed gated layers use the additional pose similarity embeddings $\hat{E}_{T \rightarrow D}^t$ as strong prior. The $\alpha$-Gates fuse the cross-attention matrix $S_A$ between detections and tracks, which measures appearance similarity, and the pose similarity matrix $S_E$, which measures spatial and pose similarity between detections and tracks. }
  
\label{fig:two-stream-decoder}
\end{figure}

\noindent 
\textbf{Attention Layer:} As baseline method, we employ a transformer decoder as proposed in \cite{transformer} and we propose an attention-based matching layer as the main extension. As illustrated in \cref{fig:two-stream-decoder}a, the attention layer calculates the similarities $A = \sigma(O_A)$ between the appearance features of the detections $E_D^t$ and tracks $E_T^{t-1}$, where $\sigma$ is the row-wise softmax layer. $O_A$ represents the appearance-based similarity logits that are obtained by cross-attention between the appearance features of the detections $E_D^t$ and tracks $E_T^{t-1}$:

\begin{equation}
    O_A=\frac{{E}_T^{t-1} W_Q^\intercal \left({E}_D^{t} W_K^\intercal\right)^\intercal}{\sqrt{d}},
\end{equation}
where $W_Q$ and $W_K$ are the learned projection weights for the queries and keys, respectively, and $d$ is the dimensionality of the embeddings $E_*$.

In order to assign detections to tracks based on appearance similarity, we need to allow that none of the detections is assigned to a track, \eg, if a person has not been detected or is occluded. Prior to applying the softmax to the attention logits, \ie, $A = \sigma(O_A)$, we thus add a column of zeros. In other words, the last column of $A$ indicates if a track does not match with any detection. 

Finally, the attention layer calculates the proposed track embedding update as follows: 
\begin{equation}\label{eq:embedding_update}
    \Delta E_{T}^t = (A_{:,:-1}E_D^t)W_A^\intercal,
\end{equation}
where $A_{:,:-1}$ denotes the attention weights without the last column and $W_A$ are the weights of a linear layer.

\noindent\textbf{Gated Attention Layer:} Appearance similarities allow to re-identify an occluded person after some frames, but are unreliable in case of motion blur or person instances with similar appearance as it is common in team sport videos. Pose and spatial similarities, on the other hand, provide a strong matching prior, but are less reliable in crowded scenarios as shown in \cref{fig:qualitative}. While none of them can resolve all ambiguities, fusing the similarities automatically provides a stronger matching prior. Based on these intuitions, we propose the gated attention layer. 
As illustrated in \cref{fig:two-stream-decoder}d, the gated attention layer extends \cref{fig:two-stream-decoder}a and incorporates pose similarity weights $S_E{=}\sigma\left(O_E\right)$ between detections and tracks. The pose similarity logits $O_E{=}E_{T\rightarrow D}^t W_E^\intercal$ are obtained by $E_{T\rightarrow D}^t$ and the learned weight matrix $W_E$.

In order to assign detections to tracks, appearance-based similarities $S_A$ and pose-based similarities $S_E$ are fused by the $\alpha{-}Gate$ that weights the contribution of appearance-based attention weights and the pose-based attention weights by a hyperparameter $\alpha$, which we evaluate in our experiments:
\begin{equation}\label{eq:weighted_attention}
    A = \alpha \cdot S_A + (1 - \alpha) \cdot S_E.
\end{equation} 
Similar to the attention layer (\cref{fig:two-stream-decoder}a), we add a column of zeros to $S_A$ and $S_E$ and obtain $\Delta E_T^t$ following \eqref{eq:embedding_update}.
Intuitively, fusing the normalized similarities $S_A$ and $S_E$ automatically assigns a higher weight to the similarity measure where the matching confidence of a detection to a track is higher, resulting in a higher tracking accuracy, as we will show in the experiments.

\vspace{-0.5cm}
\paragraph{Decoder Layer:} 

The decoder layer shown in \cref{fig:two-stream-decoder}b then updates the track $E_{T}^{t-1}$ embeddings based on the output of the attention layer. As common for transformer blocks \cite{transformer}, we use a residual feed-forward network (FNN) as shown in \cref{fig:two-stream-decoder}b. Specifically, we compute 
\begin{equation}
    \widehat{E}^t_T = LN( \widetilde{E}^{t}_T + FFN(\widetilde{E}^{t}_T)),\;\;\;\;\; 
    \widetilde{E}^{t}_T = LN(E_T^{t-1} + \Delta E_{T}^t) ,
\end{equation}
where LN denotes layer normalization. 

\noindent
\textbf{Gated Decoder Layer:} The gated decoder layer as shown in \cref{fig:two-stream-decoder}e) additionally employs an $\alpha{-}Gate$ to weight the contribution of appearance-based and pose similarity-based attention logits $O_A$ and $O_E$ similar to the gated attention layer.   We then apply a feed-forward network $FFN_E$ on the weighted sum of attention logits
\begin{equation}\label{eq:alpha2}
    \widehat{E}_{T\rightarrow D}^t = FFN_E\left(\alpha \cdot O_A + (1 - \alpha) \cdot O_E \right),
\end{equation}
 to update the pose similarity embeddings, where $FFN_E$ is a feed-forward network as $FFN$. Empirically, the gated decoder layer performs best if $\alpha$-gating is performed on the attention logits as we show in the experiments.

\paragraph{Gated Matching Layer:}
The matching layer (\cref{fig:two-stream-decoder}c) and the gated matching layer (\cref{fig:two-stream-decoder}f) comprise a structure similar to the attention layer and the gated attention layer, respectively, but differ in two aspects. 1) Both matching layers (\cref{fig:two-stream-decoder}c) and  (\cref{fig:two-stream-decoder}f) do not use the attention weights to predict a track embedding update and therefore do not consist of a linear layer after the softmax and $\alpha$-Gate, respectively. 2) The attention weights $A$ (\ie \eqref{eq:weighted_attention}) are used as assignment matrix  $\widehat{M}_{D\rightarrow T}^t$ to assign detections to tracks, as we show in \cref{fig:architecture}d. In the presence of duplicate detections, this allows to match multiple detections to a single track. Given the matching matrix $\widehat{M}_{D\rightarrow T}^t$, we utilize Hungarian matching to assign detections to tracks. All the remaining detections $i$ that have a matching probability  $\hat{m}_{i\rightarrow j} > \tau_{dup}$ for any track $j$ are considered as duplicate detections and are removed. Any other detection will initialize a new track embedding (\cref{fig:architecture}d). All tracks that have not been tracked for $\tau_{age}$ frames will be removed. We evaluate the impact of $\tau_{dup}$ and $\tau_{age}$ in the experiments.

As shown in \cref{fig:architecture}, we employ an additional \emph{Track Embedding Head} after the last decoder layer that predicts the final track embeddings and \emph{New Track Embedding Head} that generates the embedding for newly initialized tracks. Both heads consist of two linear layers, where the first layer includes a LayerNorm \cite{ba2016layer} and GELU \cite{hendrycks2016gelu}.

\subsection{Confidence-Guided Track Update}\label{p:cg_track_update} 
As shown in \cref{fig:architecture}c, we perform the final update of the tracks embedding before the final assignment of detections to tracks is carried out by the matching layer (\cref{fig:architecture}d). This is done for two reasons. Firstly, the final assignment might be wrong and, secondly, we do not want to update the track embedding when the detection embedding is noisy due to occlusion or motion blur. To prevent track embeddings from being updated by noisy or low-confident detections, we employ a confidence-guided update of the track embeddings. 
Let $\mathcal{A} = \lbrace A_{:,:-1}^n \rbrace_{n=1}^N$ be the set of dual-source attention weights of all $N$ decoder stages without the last column. As the number of detections per frame is dynamic, we apply max pooling on the rows of each attention matrix to obtain the maximum attention score for each track and estimate an importance weight $w_j$ with a linear layer as
\begin{eqnarray}
   \widehat{\mathcal{A}}_j & = & \text{concat}\lbrace \max_{i} A_{ji}^n \vert n \in [1, N]\rbrace  \\
   w_j & = & \sigma\left(\sum_n w_n \cdot \widehat{\mathcal{A}}_j^n + b_n\right),
\end{eqnarray}
where $\sigma$ is the sigmoid function and $\widehat{A}_j^n$ the maximum attention for track $j$ at layer $n$.
Ultimately, we update the embedding for each track $j$ as a confidence-guided moving average by using the importance weight $w_j$ following $E_{T_j}^t = (1 - w_j) \cdot E_{T_j}^{t-1} + w_j \cdot \widehat{E}_{T_j}^t$. As we will show in our experiments, the confidence-guided embedding update improves the performance compared to solely relying on the \emph{Track Embedding Head} (\cref{fig:architecture}c) while only adding a negligible overhead.

 \begin{table}[t]
    \begin{center}
    \begin{adjustbox}{width=\linewidth}
    \begin{tabular}{|l|ccccc|c|c|}
    \hline
    Approach & Online & AssA & FragA & DetA & HOTA & MOTA & mAP \\
        \hline\hline
    CorrTrack \cite{posetrack21} & \checkmark  & 58.02 & 57.75 & 45.48 & 51.13 & 63.0 & 72.3 \\
    CorrTrack w. ReID \cite{posetrack21} & \checkmark  &  60.21 & 59.66 & 46.56 & 52.71 & 63.8 & 72.7 \\ 
    Tracktor++ w. Poses \cite{posetrack21} & \checkmark  & 59.41 & 58.61 & 46.30 & 52.21 & 63.3 & 71.4 \\
    Tracktor++ w. Corr. \cite{posetrack21} & \checkmark  & 54.05 & 52.02 & 44.67 & 48.90 & 61.6 & \textbf{73.6} \\ 
    Ours w/o gating & \checkmark  & 44.92 & 41.96 & 45.30 & 44.82 & 52.4 & 70.2\\
    Ours w. gating & \checkmark  & \textbf{62.20} &\textbf{ 60.93 } & \textbf{47.20} & \textbf{53.94} & \textbf{64.1} & \textbf{73.6}\\
       \hline\hline
    CorrTrack \cite{posetrack21} & \xmark &  60.93 & 60.37 & 45.48 & 52.42 & 63.9 & 72.3 \\
 
    \hline 
    \end{tabular}
    \end{adjustbox}
    \end{center}
            \caption{Comparison to multi-person pose tracking methods on the PoseTrack21 dataset.}
    \label{tab:sota}
\end{table}

\subsection{Embeddings $E^t_D$ and $E_{T\rightarrow D}$}\label{sec:encoder}
We finally describe how the detection embeddings $E^t_D$ (\cref{fig:architecture}a) and pose similarity embeddings $E_{T\rightarrow D}$ (\cref{fig:architecture}b) are computed.

\paragraph{Detection Embeddings $E^t_D$:}
The embeddings that we extract from each newly detected person need to be robust to occlusion. Since the keypoint heatmaps $\mathcal{H}_D^t$ of the pose estimator provide an indicator which keypoints are occluded, we condition the embedding $E^t_D$ on $\mathcal{H}_D^t$. Specifically, we use a ResNet50 \cite{resnet} and propose Spatially Adaptive Pose DEnormalization (SPAPDE) layers, which are inspired by \cite{spade}, as an extension. Each SPAPDE layer operates on a set of keypoint heatmaps 
$\mathcal{H}_D^t \in \mathbb{R}^{N \times K \times \frac{256}{s}\times \frac{128}{s} }$ along with the current ResNet features $f\in \mathbb{R}^{N\times C\times \frac{256}{s}\times \frac{128}{s}}$ extracted from the corresponding image crops $\mathcal{I}_D^t \in \mathbb{R}^{Nx3x256x128}$, where $N$ is the number of persons in frame $t$, $s$ denotes the respective scaling factor, $K$ denotes the number of keypoints and $C$ denotes the number of feature channels.

 \begin{table}[t]
    \begin{center}
    \begin{adjustbox}{width=0.9\linewidth}
 \begin{tabular}{|l|ccccc|}
  \hline
   Method & AssA & DetA & LocA & HOTA & MOTA  \\
   \hline
    TRMOT \cite{realtimemot}& 54.98 & 40.91 & 79.92 & 46.85 & 47.2 \\
    FairMOT \cite{zhang2021fairmot} & 61.45 & 47.43 & 83.16 & 53.53 & 56.3 \\
    Tracktor++ \cite{tracktor_2019_ICCV} & 65.43 & \textbf{52.71} & \textbf{83.09} & 58.29 & \textbf{59.5} \\
    CorrTrack + ReID \cite{posetrack21}& 64.19 & 51.33 & 82.80 & 56.95 & 52.0 \\
    Ours  & \textbf{66.89} & 51.81 & 82.71 & \textbf{58.42} & 55.3 \\
   \hline
  \end{tabular}
    \end{adjustbox}
    \end{center}
  \caption{Comparison to MOT methods on the PoseTrack21-MOT dataset.}
  \label{tab:mot_results_MOTA}
\end{table}
In contrast to batch normalization \cite{10.5555/3045118.3045167}, the normalized input features are scaled and shifted with respect to the keypoint heatmaps. In particular,
SPAPDE computes the modulation parameters $\beta$ and $\gamma$ following 
\begin{equation}
    \gamma = \mathrm{conv}(a), \;\;\; \beta = \mathrm{conv}(a),\;\;a =\mathrm{ReLU}(\mathrm{conv}(\mathcal{H}_D^t)),
\end{equation}
where $\mathrm{conv}$ denotes a 3x3-convolution and $\beta, \gamma \in \mathbb{R}^{N\times C\times \frac{256}{s}\times \frac{128}{s}}$.
The image features $f$ are then conditioned on the heatmaps $\mathcal{H}_D^t$ as follows. Let $f_{n,c,y,x}$ be the feature value for the detected person $n$ and feature channel $c$ at pixel location $(x,y)$. The SPADPE layer first calculates the mean $\mu_c$ and standard deviation $\sigma_c$ over all persons and pixels of $f$ and then adaptively de-normalizes the image features by 
\begin{equation}\label{eq:spapde}
    \widehat{f}_{n,c,y,x} = \gamma_{n, c,y,x} \cdot \frac{f_{n,c,y,x} - \mu_c}{\sigma_c} + \beta_{n,c,y,x}.
\end{equation}
We replace every batch normalization layer within ResNet50 by SPAPDE layers and train the network following \cite{bagoftricks}.

Moreover, we use a vanilla N-stage transformer encoder \cite{transformer} without positional encoding to further disentangle the backbone features of severely occluded persons and noisy pose predictions and generate a set of encoded person features $E_D^T \in \mathcal{R}^{N\times 256}$ as shown in \cref{fig:architecture}a.

\vspace{-0.4cm} \paragraph{Pose Similarity Embeddings $E_{T\rightarrow D}$:} The pose similarity features $E_{T\rightarrow D}^t$ (\cref{fig:architecture}b) used in our gated attention architecture are based on similarities between bounding boxes and poses using Intersection over Union (IoU) and Object Keypoint Similarity (OKS), respectively.
 In this work, we rely on three variants of OKS: the first variant considers keypoints which are present in both poses. While the first variant provides a good measure of keypoint alignment, its expressiveness suffers if two poses only share a small subset of keypoints. 
 For that reason, the remaining two variants consider all keypoints present in one of the two poses, respectively. 
In order to deal with motion, we use as in \cite{rafi2020selfsupervised} a warping function $\phi_W$ that warps the last observed pose of all tracks into the current frame $t$. The pose similarity $E_{T\rightarrow D}$ between tracks and detections is then computed as follows:
\begin{equation}
    E_{T\rightarrow D} = \phi_E\left(\left[IOU(\widehat{\mathcal{B}}_T^{t-1},\mathcal{B}_D^{t})\;||\;OKS(\widehat{\mathcal{P}}_T^{t-1},\mathcal{P}_D^{t})\right]\right).
\end{equation}
Here, $\widehat{\mathcal{P}}_T^{t-1} = \phi_W(\mathcal{P}_T^{t-1})$ and $\widehat{\mathcal{B}}_T^{t-1} = \phi_W(\mathcal{B}_T^{t-1})$ represent the set of warped track poses and track bounding boxes, respectively. The operator $\left[ \cdot\;||\;\cdot \right]$ represents concatenation, and $\phi_E$ denotes the pose similarity embedding head, which consists of three linear layers with LayerNorm \cite{ba2016layer} and GELU \cite{hendrycks2016gelu}.

 \begin{table}[t]
    \begin{center}
    \begin{adjustbox}{width=0.95\linewidth}
    \begin{tabular}{|l|c|c|c|cc|}
    \hline
    Approach & Online & Val. Set & Detector & MOTA & mAP \\
    \hline\hline
    STAF \cite{STAF} & \checkmark & v1 & - & 60.9 & 70.4 \\
    T CPN++ \cite{10.1007/978-3-030-11012-3_19} & \checkmark & v1 & Cascade R-CNN \cite{cai2019cascadercnn} & 64.0 & 80.9\\
    MIPAL \cite{mipal} & \checkmark & v1 & - & 65.7 & 74.6\\
    KeyTrack \cite{15_keypoints} & \checkmark & v1 & HTC \cite{Chen2019HybridTC}  & 66.6 & 81.6 \\
    CorrTrack \cite{rafi2020selfsupervised} & \checkmark &  v1 & Cascade R-CNN \cite{cai2019cascadercnn} & 68.8 & 79.2 \\
    TKMRNet \cite{zhou2020temporal} & \checkmark &  v1 & Faster R-CNN FPN DCN \cite{zhu2018deformable} &  68.9 & 76.7 \\
 
    CorrTrack \cite{rafi2020selfsupervised} & \xmark & v1 & Cascade R-CNN \cite{cai2019cascadercnn} & 69.1 & 79.2  \\
   
    \hline\hline
    CorrTrack \cite{rafi2020selfsupervised} & \checkmark & v2 & Cascade R-CNN \cite{cai2019cascadercnn} &  63.6 & 75.9 \\ 
  
    LTIVA \cite{learning_to_track_instances} & \checkmark & v2 & - & 64.7 & 71.4 \\
  
    CombDet \cite{combining_detection_and_tracking} & \xmark & v2 & ResNet-101 SNIPER \cite{Singh2018SNIPEREM} &  68.7 & 81.5 \\
  
     LDGNN \cite{learning_dynamics_via_graph} & \checkmark & v2 & Faster R-CNN FPN DCN \cite{zhu2018deformable} &  69.2 &  77.9 \\
     \hline 
    Ours & \checkmark & v2 & Cascade R-CNN \cite{cai2019cascadercnn} &  64.5 & 76.4 \\

    \hline
  
    \hline 
    \end{tabular}
    \end{adjustbox}
    \end{center}
            \caption{Comparison to the state of the art on PoseTrack 2018 \cite{posetrack18}. Two versions of the validation set have been released containing 74 (v1) and 170 (v2) sequences, respectively.}
    \label{tab:sota_pt18}
    \vspace{-2mm}
\end{table}

\subsection{Training Objective}\label{sec:train_objective}
In a first step, we train the re-identification network following \cite{bagoftricks}: We apply the triplet loss \cite{HermansBeyer2017Arxiv} and center loss \cite{Wei2017GLADGD} after the last pooling layer of ResNet50 and we employ the cross-entropy loss with label smoothing \cite{7780677} on the classification layer. Subsequently, we freeze the re-identification network and proceed to train our network.

In our approach, the matching layer is trained using a cross-entropy loss, which is defined as follows:
\begin{equation}\label{eq:l_match}
 \mathcal{L}_{match} = -\frac{1}{N_D} \sum_i y_i\cdot log(p^m_{ij}) + (1-y_i) \cdot p^m_{i0},
\end{equation}
where $N_D$ is the total number of detections, $p^m_{i,j}$ represents the probability of matching the $i$-th detection to its corresponding ground truth track $j$, and $p^m_{i0}$ is the probability of not matching the $i$-th detection to any track. The variable $y_i$ takes the value 1 if the $i$-th detection is assigned to a ground truth track and 0 otherwise.

Since we discard duplicates after the final matching layer, we generate duplicates during training and allow multiple detection assignments to a single track. 
Specifically, we use detected poses and ground truth poses in the training process that can share the same person identity. To assign identities to the detected poses during training, we employ OKS-based greedy matching to the ground truth poses. 

We then utilize a duplicates-aware cross-entropy loss function that operates on the attention weights of the encoder and decoder layers. The loss function is defined as follows:
\small
\begin{equation}\label{eq:l_attn}
\mathcal{L}_{attn} = -\frac{1}{N_T}\sum_j \log(p_j),\;\;\; p_j = \left(\sum_i A_{ji} \mathbb{I}_{i}(j)\right) + \mathbb{I}_{\nexists i}(j)A_{j0}.   
\end{equation}
\normalsize
$N_T$ represents the total number of tracks and $p_j$ denotes the accumulated matching probability for track $j$,

where $A_{ji}$ is the attention weight of the respective encoder/decoder layer and 
$\mathbb{I}_{i}(j)$
is 1 if the identity of the current track $j$ and the detection $i$ are the same, and 0 otherwise. If none of the detections matches, \ie, $\mathbb{I}_{\nexists i}(j)$, 
we maximize the no-match probability $A_{j0}$, which is the last column of the attention matrix as discussed in \cref{sec:twostream}. In other words, we want that $A_{ji}$ is large for the correct assignment if and only if a match exists.    

The final objective function is a combination of the cross-entropy loss function for the matching layer ($\mathcal{L}_{match}$) and the duplicate-aware cross-entropy loss functions for each encoder and decoder layer:
\begin{equation}\label{eq:L}
    \mathcal{L} = \mathcal{L}_{match} + \sum_k \mathcal{L}_{attn}^{enc_k} + \sum_k \mathcal{L}_{attn}^{dec_k},
\end{equation}
where $\mathcal{L}_{attn}^{enc_k}$ and $\mathcal{L}_{attn}^{dec_k}$ represent the duplicate-aware cross-entropy loss functions for the $k$-th encoder and decoder layer, respectively.

\section{Experiments}

 \begin{table}[t]
   \begin{center}
    \begin{adjustbox}{width=0.8\linewidth}
    \begin{tabular}{|cccc|c|}
    \hline
     $\mathcal{L}_{attn}^{dec}$ & Encoder & $\mathcal{L}_{attn}^{enc}$ & CG-Update & HOTA \\
     \hline
      & &  &  & 53.04  \\
      \checkmark & &  &  & 53.40  \\
      \checkmark & \checkmark &  &  & 53.56  \\

      \checkmark & \checkmark & \checkmark &  & 53.61  \\
      \checkmark & \checkmark & \checkmark & \checkmark & 53.94  \\
    
     \hline
    \end{tabular}
    \end{adjustbox}
    \end{center}
    \caption{Impact of several components in our pose tracking network on the tracking performance. CG-Update denotes the Confidence-Guided Track Update as discussed in \cref{sec:twostream}.}
    \label{tab:architecture_search}
\end{table}
\subsection{Datasets and Evaluation}
We evaluate our work on the PoseTrack datasets \cite{posetrack18,posetrack21}. Both datasets are large-scale benchmarks for multi-person pose tracking and contain 593 videos for training and 170 for evaluation. The videos contain various activities and include highly diverse poses and severe occlusions as shown in \cref{fig:qualitative}. Since for PoseTrack 2018 the evaluation server is not anymore available, we only report results on the validation set. Compared to PoseTrack 2018, PoseTrack21 \cite{posetrack21} provides more annotations and additional benchmarks for multi-object tracking (MOT) and person search. We thus primarily focus on PoseTrack21 \cite{posetrack21} in our experiments.
For evaluation, we use \textit{keypoint HOTA} \cite{posetrack21}. Keypoint HOTA consists of sub-metrics that measure the \textit{detection accuracy (DetA)}, the \textit{association accuracy (AssA)} and the \textit{fragmentation accuracy (FragA)}. In addition, we report results for the \textit{keypoint-based MOTA} metric \cite{posetrack18}. Both metrics are evaluated on a keypoint level and then averaged. For completeness, we report the keypoint detection performances in terms of \textit{mean average precision (mAP)}. 

We follow common practice \cite{15_keypoints,simple_baseline,rafi2020selfsupervised,learning_dynamics_via_graph,posetrack21,combining_detection_and_tracking} and utilize a multi-frame pose estimation approach to compensate for missed detections due to motion blur and occlusions during inference. In particular, we utilize keypoint correspondences as in \cite{rafi2020selfsupervised,posetrack21}. In the following, we compare our approach to the state of the art. Implementation details and additional ablation studies are provided as supplementary material.

\subsection{Comparison with State of the Art}

 \begin{table}[t]
   \begin{center}
    \begin{adjustbox}{width=0.8\linewidth}
    \begin{tabular}{|l|c|c|}
    \hline
     Re-ID Network & Pose-Conditioned & HOTA \\
     \hline\hline
     ResNet50~\cite{bagoftricks}  &  & 53.47 \\
     SPAPDE (ResNet50)  & \checkmark &  53.94\\
     \hline
    \end{tabular}
    \end{adjustbox}
    \end{center}
    \caption{Impact of the proposed SPAPDE network for re-identification on the overall performance on PoseTrack21.}
    \label{tab:diff_reid_backbone}
\end{table}

\noindent\textbf{PoseTrack21:}
We first evaluate our model with and without gating on the PoseTrack21 validation set and compare the performance to methods proposed in \cite{posetrack21} using the keypoint HOTA \cite{posetrack21} and the MOTA metrics \cite{posetrack18}. The results are shown in \cref{tab:sota}. While the performance \textit{without} gated attention is quite low, our proposed gated attention transformer  consistently outperforms existing methods, achieving a HOTA score of 53.94 and a MOTA score of 64.1.  Compared to \textsc{CorrTrack W. ReID}, our approach boosts the association accuracy (AssA) and fragmentation accuracy (FragA) by $+1.99\%$ and $+1.27\%$ to $62.20$ and $60.93$, respectively. Additionally, the detection accuracy (DetA) and the mAP increase to $47.20$ and $73.6$, respectively. While our approach performs online multi-person pose tracking, it also outperforms the offline approach CorrTrack.

We further evaluate our approach on the PoseTrack21-MOT benchmark and compare the performance to the methods in \cite{posetrack21}. As we show in \cref{tab:mot_results_MOTA}, our approach consistently outperforms existing methods in terms of AssA (66.89) and HOTA (58.42). 
Tracktor++~\cite{tracktor_2019_ICCV} achieves a slightly higher DetA and localization accuracy (LocA), which also results in higher MOTA. While Tracktor++ has been trained on the annotated bounding boxes for MOT, our approach has been trained for pose tracking and we simply generate the bounding boxes from the estimated poses. Consequently, MOT methods achieve a better MOTA score due to higher bounding box detection and localization accuracy. As discussed in \cite{luiten2020IJCV}, HOTA is a better metric than MOTA for MOT. 

\noindent\textbf{PoseTrack 2018:} The comparison with related works on PoseTrack 2018 is difficult for two reasons: i) The PoseTrack 2018 dataset is not available anymore and it is no longer possible to submit results to the official test server; ii) the validation set was released in two different versions. The first version (v1) contains 74 whereas the second version (v2) contains 170 sequences, respectively. 
For completeness, we also include results that have been reported for v1 in \cref{tab:sota_pt18}, but these numbers are not comparable. The results for CorrTrack \cite{rafi2020selfsupervised} show that the version v2 is much more difficult. 
Compared to CorrTrack, our method improves the MOTA score by $0.9\%$ to $64.5$. 
The pose estimation performance increases from $75.9$ to $76.4$ in terms of mAP. Our appraoch performs similar to \cite{learning_to_track_instances} in terms of MOTA while achieving a higher mAP. CombDet \cite{combining_detection_and_tracking} achieves a higher accuracy, but it uses a stronger multi-frame person detector and is an offline approach, whereas our approach is an online approach. LDGNN \cite{learning_dynamics_via_graph} also uses a better multi-frame pose estimator, but the code of the pose estimator is not publicly available.

\begin{table}[t]
    \begin{center}
    \begin{adjustbox}{width=\linewidth}
    \begin{tabular}{|c|c|cccc|}
    \hline
     $\alpha$-Gate Source & Learnable $\alpha$ &  AssA & FragA & DetA & HOTA \\
    \hline
    Attention Logits \; $O_A$ \& $O_E$ & & 62.20 & 60.93 & 47.20 & 53.94 \\ 
    Attention Logits \; $O_A$ \& $O_E$ & \checkmark & 61.80 & 60.59 & 47.15 & 53.74 \\
    Attention Weights $S_A$ \& $S_E$ & & 54.74 & 50.77 & 47.14 & 50.53 \\
   
    \hline
    
    \end{tabular}
    \end{adjustbox}
    \end{center}
            \caption{Impact of using attention weights or attention logits in the $\alpha$-Gate \eqref{eq:alpha2} of the Gated Attention Transformer on the overall performance on PoseTrack21.}

    \label{tab:additional_ablations}
\end{table}

\begin{figure*}[t]
\begin{center}
   \includegraphics[width=\textwidth]{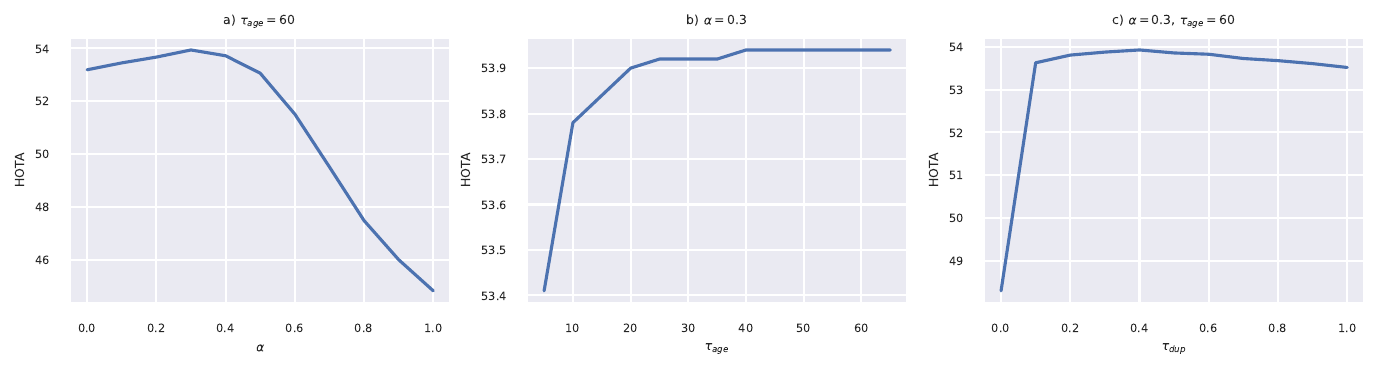}
\end{center}
   \caption{Impact of the parameters $\alpha$, maximum track ages $\tau_{age}$ and the duplicate confidence threshold $\tau_{dup}$ on the tracking performance. a) visualizes the impact of $\alpha$ with $\tau_{age}=60$ and $\tau_{dup} = 0.4$. b) shows the performance evaluation with respect to the maximum track age $\tau_{age}$ with $\alpha=0.3$ and $\tau_{dup}=0.4$. c) illustrates the impact of $\tau_{dup}$ with $\alpha = 0.3$ and $\tau_{age}=60$.   }
\label{fig:xval_alpha_track_age}
\end{figure*}

\subsection{Ablation Studies}
\subsubsection{Evaluation of the Network Architecture}
We perform ablation experiments to examine the influence of each building block of our proposed method. All experiments are conducted on the PoseTrack21 dataset. 
\vspace{-0.3cm}\paragraph{Loss Terms, Transformer Encoder and Confidence-Guided Track Update:}
We first evaluate the impact of each component in our tracking model on the keypoint HOTA score \cite{posetrack21} in \cref{tab:architecture_search}. We start with the gated attention decoder with matching layer as our base model, which we trained with the matching loss $\mathcal{L}_{match}$~\eqref{eq:l_match}. Subsequently, we incrementally activate several components and evaluate their impact on the overall performance. Direct supervision on the gated attention layers using $\mathcal{L}_{attn}^{dec}$ \eqref{eq:l_attn} increases the overall performance from $53.04$ to $53.40$ in terms of HOTA. 
Using additional transformer encoder layers (\cref{fig:architecture}a) boosts the performance to $53.56$. Using the loss $\mathcal{L}_{attn}^{enc}$ \eqref{eq:l_attn} also for the encoder increases HOTA to $53.61$. The impact of the Confidence-Guided Track Update (\cref{p:cg_track_update}) is shown in the last row. It further increases the tracking performance by $0.33$ to a HOTA score of $53.94$. 

\vspace{-0.2cm}
\paragraph{Pose Similarity Embedding Update:}

To update the pose similarity embeddings in the gated decoder layer, we utilize an $\alpha$-Gate to weight the contributions of the appearance-based and pose similarity-based attention logits \eqref{eq:alpha2}. While we evaluate the impact of $\alpha$ in Fig.~\ref{fig:xval_alpha_track_age}a, \cref{tab:additional_ablations} shows that learning $\alpha$ does not improve the results.  
We further evaluate the tracking performance when the fusion is not done at the logits, \ie, before the softmax, but after the softmax as in \eqref{eq:embedding_update}, \ie,    
\begin{equation}\label{eq:alpha3}
    \widehat{E}_{T\rightarrow D}^t = FFN_E\left(A_{:,:-1} \right),
\end{equation}
where $A_{:,:-1}$ denotes the attention weights without the last column (\cref{sec:twostream}). The last row in \cref{tab:additional_ablations} shows that the performance largely decreases from a HOTA score of $53.94$ to $50.53$ in this case.

\vspace{-0.4cm}
\paragraph{Re-Identification Model:}
To evaluate the impact of the re-identification network, we trained our tracking model with two different re-identification networks, which we previously trained on PoseTrack21. We used the re-identification network from \cite{bagoftricks} and the proposed pose-conditioned SPAPDE network (\cref{sec:encoder}). \cref{tab:diff_reid_backbone} shows that adding Spatially Adaptive Pose DE-normalization layers (SPAPDE) to the network increases HOTA from $53.47$ to $53.94$.

\subsubsection{Hyperparameter Evaluation}

If not otherwise specified, we use in all experiments $\alpha=0.3$, $\tau_{age}=60$ and $\tau_{dup}=0.4$. We finally evaluate the impact of these parameters on the PoseTrack21 dataset.  

\vspace{-0.4cm}\paragraph{Impact of $\alpha$:}
We evaluate the impact of $\alpha$ in the $\alpha$-Gate, \eqref{eq:weighted_attention} and \eqref{eq:alpha2}, in \cref{fig:xval_alpha_track_age}a. $\alpha$ weights the contribution of the appearance-based attention weights and the pose-based attention weights, where $\alpha=1.0$ only considers appearance-based attention weights and, vice versa, $\alpha=0.0$ only considers pose similarity-based attention weights. As we can observe in \cref{fig:xval_alpha_track_age}a, the tracking performance drastically decreases for $\alpha > 0.4$. Appearance-based person features are very sensitive to persons with similar appearance as it is common in sports videos, resulting in false associations and a high degree of identity switches. For $0 \leq \alpha \leq 0.3$, we can observe a linear increase in the overall performance, peaking at $\alpha=0.3$.
This shows that pose-based and appearance-based similarities complement each other. Pose similarities provide strong guidance between consecutive frames, while appearance-based features allow to recover inactive tracks, \eg, due to occlusion.

\vspace{-0.4cm}\paragraph{Impact of $\tau_{age}$:} We close tracks that have not been tracked for more than $\tau_{age}$ frames and do not include them for the detection-to-track matching anymore. \cref{fig:xval_alpha_track_age}b shows that the accuracy saturates at $\tau_{age}=40$. 

\vspace{-0.4cm}\paragraph{Impact of $\tau_{dup}$:}
During tracking, we remove unmatched detections if they have a matching confidence $m_{ij} > \tau_{dup}$ with an already matched track (\cref{sec:twostream}). \cref{fig:xval_alpha_track_age}c shows that the accuracy drops without such a threshold since duplicates generate new tracks in this case. 

\section{Conclusion}
We presented a novel gated attention approach for multi-person pose tracking. Our method employs a duplicate-aware association and dynamically adapts via gates the impact of pose-based similarities and appearance-based similarities based on the attention probabilities of each similarity measure. We evaluated our approach on the challenging PoseTrack21 dataset where our approach outperforms previous works for multi-person pose tracking. On PoseTrack 2018, the approach is only outperformed by methods that use a more expensive human pose estimator. We also evaluated the impact of the proposed Spatially Adaptive Pose DEnormalization (SPAPDE) on the pose tracking performance on PoseTrack21, which positively impacts the overall tracking performance.

\vspace{-0.4cm}\paragraph{Acknowledgements} This work has been funded by the Deutsche Forschungsgemeinschaft (DFG, German Research Foundation) - GA 1927/8-1.

\appendix
\section{Implementation Details}\label{app:implementation_detail}
\begin{figure}[t]
\begin{center}
\includegraphics[width=\linewidth]{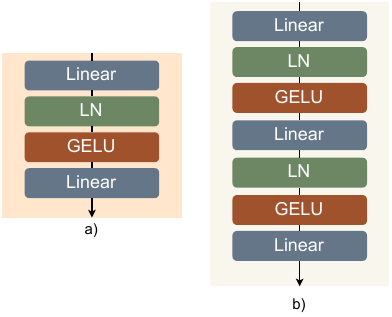}
\vspace{-5mm}
\end{center}
   \caption{Illustration of the a) track embedding head and b) temporal pose similarity embedding head. }
\label{fig:embedding_head}
\end{figure}

\paragraph{Track Embedding and Pose Similarity Embedding Heads:}
\cref{fig:embedding_head} visualizes the structure of the proposed \textit{Track Embedding Head}  (\cref{fig:embedding_head}a) and the \textit{Temporal Pose Similarity Embedding Head} (\cref{fig:embedding_head}b). The \textit{New Track Embedding Head} (\cref{fig:architecture}d)) has the same structure as the \textit{Track Embedding Head}.

\paragraph{Re-Identification Model:} We train the re-identification model for 244 epochs on the PoseTrack21 person search dataset with a batch size of 256 and a learning rate of $0.00035$ that we decay to $0.000035$ after 75 epochs. During the first 10 epochs, we apply a linear learning rate warm-up. Additionally, we apply data augmentation such as random scaling, random rotation and horizontal flipping.

\paragraph{Gated Attention Transformer:} Our proposed gated attention transformer employs two encoder and two decoder stages. On the PoseTrack21 dataset, we train our transformer for 14 epochs with a learning rate of $0.0001$, which is decayed by a factor of 10 after 13 epochs. We further incorporate a linear learning rate warm-up over the course of 16k iterations and optimize the network with the AdamW \cite{adamw} optimizer. Each training sequence is split into sub-sequences of length three with an overlap of one frame.
We follow the same settings on the PoseTrack 2018 dataset and train the gated attention transformer for 11 epochs. 

\paragraph{Person Detector:} For a fair comparison to related works on PoseTrack21, we utilize the same person detector and pose estimation model for all our experiments. In particular, we use the person detector from \cite{posetrack21}, which consists of a FasterRCNN \cite{fasterRCNN} with a ResNet50-FPN \cite{fpn} backbone. The detector was first pre-trained on MSCOCO \cite{mscoco} and further fine-tuned on PoseTrack21 for 30 more epochs. As pose estimator, we employ the released model \cite{posetrack21}, that was originally proposed in \cite{rafi2020selfsupervised}. The pose estimation model was trained on MSCOCO and PoseTrack21 for 215 and 16 epochs, respectively. On PoseTrack 2018, we use Cascade R-CNN \cite{cai2019cascadercnn} as object detector and the pose estimator from CorrTrack \cite{rafi2020selfsupervised}. 

\section{Ablation Studies}
\noindent We perform additional ablation experiments to examine the influence of various building blocks of our proposed method. All experiments are conducted on the PoseTrack21 dataset. 
\subsection{Evaluation of the Network Architecture}
\paragraph{New Track Embedding Head:}
Given detections in frame $t$ that can not be matched to any existing track, we use the \textit{New Track Embedding Head} (\cref{fig:architecture}d)) to generate new track embeddings from the unmatched detection embeddings. In \cref{tab:additional_head}, we evaluate the impact of the \emph{New Track Embedding Head} compared to initializing new tracks directly from the detection embeddings. The \emph{New Track Embedding Head} provides a better initialization of the track embeddings.  

\begin{table*}[t!]
   \begin{center}
    \begin{adjustbox}{width=0.9\textwidth}
    \begin{tabular}{|l|c|c|c|c|c|c|}
    \hline
        Model & GT Poses & Kernel Width & Sampling & pre-trained & mAP & loss\\
        \hline\hline
        ResNet50 & & &  &  \checkmark & 64.24 & triplet\\
        ResNet50 & & & \checkmark &  \checkmark & 68.31 & triplet\\
        ResNet50 & & & \checkmark & & 68.65 & triplet\\
        ResNet50 \& back-of-tricks \cite{bagoftricks}  &&& \checkmark & & 73.38 & triplet + ce + center\\
        
        \hline
        SPAPDE ResNet50 &\checkmark& 2& \checkmark && 66.77 & triplet \\
        SPAPDE ResNet50 &\checkmark& 5   &   \checkmark  && 70.39 & triplet\\
        SPAPDE ResNet50 &\checkmark& 10 & \checkmark & & 71.17 & triplet\\
        SPAPDE ResNet50 &\checkmark& 15 & \checkmark & & 69.62 & triplet\\
        \hline 
     
        SPAPDE ResNet50 \& bag-of-tricks \cite{bagoftricks}& \checkmark &  10 & \checkmark && \textbf{78.00} & triplet + ce + center \\
        SPAPDE ResNet50 \& bag-of-tricks \cite{bagoftricks} & & 10 & \checkmark && \textbf{74.42} & triplet + ce + center \\
        \hline 
    \end{tabular}
    \end{adjustbox}
    \end{center}
    \caption{Person re-identification performance (mAP) on the PoseTrack21 dataset for various model settings.}
    \label{tab:reid_ablation}
\end{table*}

\paragraph{Person Re-Identification}
We evaluate the person re-identification model on PoseTrack21 and measure the performance in terms of mean average precision (mAP), \ie, we calculate the area under the Precision-Recall curve for each query and average over all queries.

As we show in \cref{tab:reid_ablation}, training a ResNet50 \cite{resnet} with proper data sampling results in a significant performance gain and results in a mAP score of $68.65$. In more detail, we sample $K=6$ different instances of the same person identity for every batch. Surprisingly, training the ResNet50 from scratch results in a better mAP performance ($68.65$)  compared to training a ResNet50 pre-trained on ImageNet \cite{deng2009imagenet} ($68.31$). As discussed in \cref{sec:encoder}, we follow \cite{bagoftricks}, which further increases the performance to $73.78$.
To incorporate pose information, we replace each batch normalization layer by SPAPDE layers (\eqref{eq:spapde}), which incorporates pose information by the keypoint heatmaps. The best performance is achieved by using keypoint heatmaps with a kernel size of 10. Specifically, we denote the kernel size as the standard deviation of a Gaussian distribution. For each keypoint we calculate a Gaussian distribution with the mean set to the respective keypoint location and a standard deviation of 10.
Further following \cite{bagoftricks}, we achieve a total performance of $78.0$ and $74.42$ with ground truth and estimated poses, respectively. By adding SPAPDE, mAP thus increases from $73.78$ to $74.42$.

\paragraph{Impact of different pose similarity embeddings}
\cref{tab:diff_person_sim} shows the impact of different embedded pose similarities. IoU only provides coarse information and does not allow to distinguish between spatially closely located person instances. Even though a temporal similarity based on IoU achieves a HOTA score of $53.59$ and outperforms all related works (\cref{tab:sota}), OKS-based temporal similarity performs better and the combination performs best.

\subsection{Runtime Comparison}

Finally, we measure the average runtime of our proposed Gated Attention Transformer in comparison to the baseline method \textsc{CorrTrack + ReID}. As both methods are not limited to a specific pose estimation framework, we estimate the tracking runtime independent of the person detector and pose estimation pipelines. While the pose estimation pipeline runs at 2.3 frames per second (fps), the tracking stage of \textsc{CorrTrack + ReID} achieves an average runtime of 6.54 fps. In contrast, our proposed tracker is \textbf{3.8} times faster and runs with an average runtime of 25.07 fps.

\section{Qualitative Results}
\begin{figure*}
    \centering

    \subfloat[]{ \includegraphics[width=\linewidth,height=5cm]{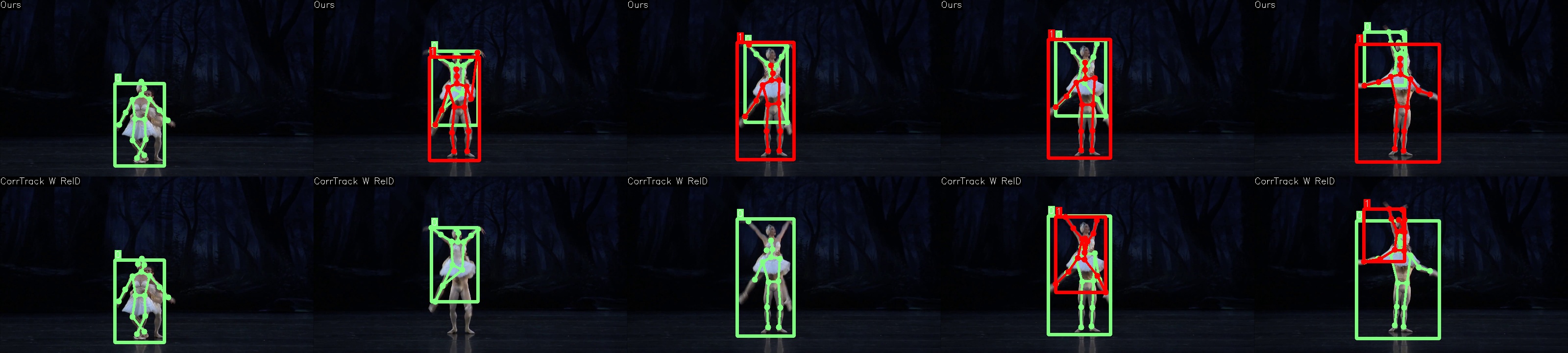}} \\
    \subfloat[]{ \includegraphics[width=\linewidth,height=5cm]{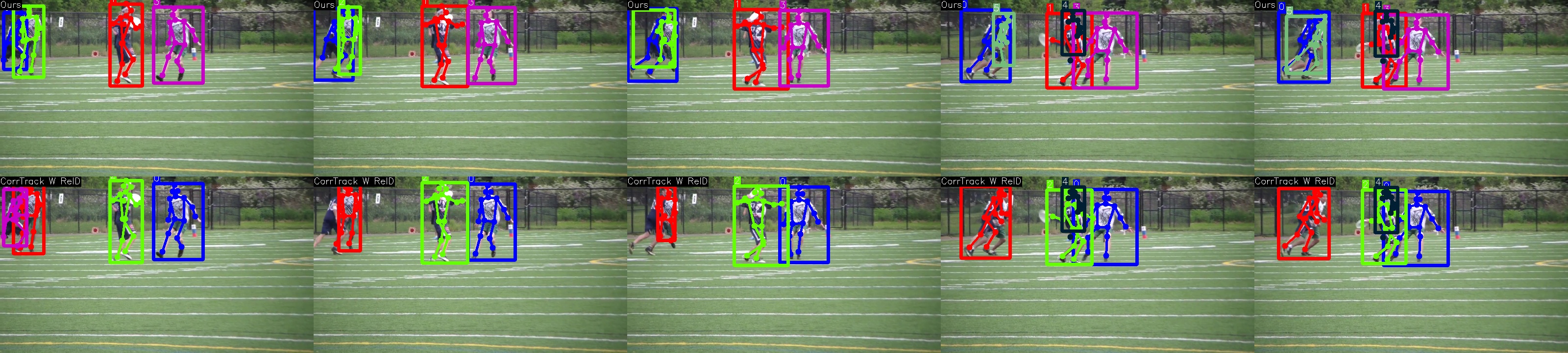}}

    \hfill
    \caption{Additional qualitative examples of our proposed method on the PoseTrack21 dataset. In both (a) and (b), the first row contains visual tracking results of our method and the second row shows visualizations of CorrTrack with ReID \cite{posetrack21}.}
    \label{fig:qualitative_2}
\end{figure*}

Fig.~\ref{fig:qualitative_2} shows some additional qualitative results. Videos are available at      
\href{https://youtu.be/uFXD4mWPajo}{https://youtu.be/uFXD4mWPajo} and
\href{https://youtu.be/lG6GkWSlNQU}{https://youtu.be/lG6GkWSlNQU}.

\begin{table}[t]
    \begin{center}
    \begin{adjustbox}{width=\linewidth}
    \begin{tabular}{|c|cccc|}
    \hline
    Track Embedding Initialization &  AssA & FragA & DetA & HOTA \\
    \hline
    Detection Embedding & 61.87 & 60.73 & 47.15 & 53.76 \\
    New Track Embedding Head & 62.20 & 60.93 & 47.20 & 53.94 \\ 
   
    \hline
    
    \end{tabular}
    \end{adjustbox}
    \end{center}
            \caption{Impact of the \textit{New Track Embedding Head} on the overall tracking performance on PoseTrack21.}
    \label{tab:additional_head}
\end{table}

 \begin{table}[t]
   \begin{center}
    \begin{adjustbox}{width=0.6\linewidth}
    \begin{tabular}{|c|c|}
    \hline
     Temporal Person Similarity & HOTA \\
     \hline\hline
      IoU & 53.59\\
      OKS & 53.81\\
      IoU + OKS & 53.94\\
     \hline
    \end{tabular}
    \end{adjustbox}
    \end{center}
    \caption{Impact of different temporal person similarities used for the calculation of the pose similarity embeddings. }
    \label{tab:diff_person_sim}
\end{table}

{\small
\bibliographystyle{ieee_fullname}
\bibliography{egbib}

\begin{thebibliography}{10}\itemsep=-1pt

\bibitem{posetrack18}
Mykhaylo Andriluka, Umar Iqbal, Anton Milan, Eldar Insafutdinov, Leonid
  Pishchulin, Juergen Gall, and Bernt Schiele.
\newblock {PoseTrack: A Benchmark for Human Pose Estimation and Tracking}.
\newblock In {\em CVPR}, 2018.

\bibitem{ba2016layer}
Jimmy~Lei Ba, Jamie~Ryan Kiros, and Geoffrey~E. Hinton.
\newblock {Layer Normalization}, 2016.

\bibitem{tracktor_2019_ICCV}
Philipp Bergmann, Tim Meinhardt, and Laura Leal{-}Taix{\'{e}}.
\newblock {Tracking without bells and whistles}.
\newblock In {\em ICCV}, 2019.

\bibitem{cai2019cascadercnn}
Zhaowei Cai and Nuno Vasconcelos.
\newblock {Cascade R-CNN: High Quality Object Detection and Instance
  Segmentation}.
\newblock {\em arXiv preprint arXiv:1906.09756}, 2019.

\bibitem{Chen2021KeypointMP}
Di Chen, Andreas Doering, Shanshan Zhang, Jian Yang, Juergen Gall, and Bernt
  Schiele.
\newblock Keypoint message passing for video-based person re-identification.
\newblock In {\em AAAI}, 2021.

\bibitem{Chen2018VideoPR}
Dapeng Chen, Hongsheng Li, Tong Xiao, Shuai Yi, and Xiaogang Wang.
\newblock Video person re-identification with competitive snippet-similarity
  aggregation and co-attentive snippet embedding.
\newblock {\em CVPR}, 2018.

\bibitem{Chen2019HybridTC}
Kai Chen, Jiangmiao Pang, Jiaqi Wang, Yu Xiong, Xiaoxiao Li, Shuyang Sun,
  Wansen Feng, Ziwei Liu, Jianping Shi, Wanli Ouyang, Chen~Change Loy, and
  Dahua Lin.
\newblock {Hybrid Task Cascade for Instance Segmentation}.
\newblock {\em CVPR}, 2019.

\bibitem{chen2023beyond}
Weihua Chen, Xianzhe Xu, Jian Jia, Hao Luo, Yaohua Wang, Fan Wang, Rong Jin,
  and Xiuyu Sun.
\newblock Beyond appearance: a semantic controllable self-supervised learning
  framework for human-centric visual tasks.
\newblock In {\em CVPR}, 2023.

\bibitem{deng2009imagenet}
Jia Deng, Wei Dong, Richard Socher, Li-Jia Li, Kai Li, and Li Fei-Fei.
\newblock {Imagenet: A large-scale hierarchical image database}.
\newblock In {\em CVPR}, 2009.

\bibitem{Dijkstra1959}
E.~W. Dijkstra.
\newblock {A note on two problems in connexion with graphs}.
\newblock {\em Numerische Mathematik}, 1959.

\bibitem{posetrack21}
Andreas Doering, Di Chen, Shanshan Zhang, Bernt Schiele, and Juergen Gall.
\newblock {PoseTrack21: A Dataset for Person Search, Multi-Object Tracking and
  Multi-Person Pose Tracking}.
\newblock In {\em CVPR}, 2022.

\bibitem{doering18}
Andreas Doering, Umar Iqbal, and Juergen Gall.
\newblock {Joint Flow: Temporal Flow Fields for Multi Person Tracking}.
\newblock {\em CVPR}, 2018.

\bibitem{vit}
Alexey Dosovitskiy, Lucas Beyer, Alexander Kolesnikov, Dirk Weissenborn,
  Xiaohua Zhai, Thomas Unterthiner, Mostafa Dehghani, Matthias Minderer, Georg
  Heigold, Sylvain Gelly, Jakob Uszkoreit, and Neil Houlsby.
\newblock {An Image is Worth 16x16 Words: Transformers for Image Recognition at
  Scale}.
\newblock In {\em ICCV}, 2021.

\bibitem{learning_to_track_instances}
Yang Fu, Sifei Liu, Umar Iqbal, Shalini~De Mello, Humphrey Shi, and Jan Kautz.
\newblock {Learning to Track Instances without Video Annotations}.
\newblock {\em CVPR}, 2021.

\bibitem{gao2020pose}
Shang Gao, Jingya Wang, Huchuan Lu, and Zimo Liu.
\newblock {Pose-guided Visible Part Matching for Occluded Person ReID}.
\newblock In {\em CVPR}, 2020.

\bibitem{detect_n_track}
Rohit Girdhar, Georgia Gkioxari, Lorenzo Torresani, Manohar Paluri, and Du
  Tran.
\newblock {Detect-and-Track: Efficient Pose Estimation in Videos}.
\newblock In {\em CVPR}, 2018.

\bibitem{resnet}
Kaiming He, Xiangyu Zhang, Shaoqing Ren, and Jian Sun.
\newblock {Deep Residual Learning for Image Recognition}.
\newblock In {\em CVPR}, 2016.

\bibitem{hendrycks2016gelu}
Dan Hendrycks and Kevin Gimpel.
\newblock {Gaussian Error Linear Units (GELUs)}.
\newblock {\em arXiv-Preprint}, 2016.

\bibitem{HermansBeyer2017Arxiv}
Alexander Hermans, Lucas Beyer, and Bastian Leibe.
\newblock {In Defense of the Triplet Loss for Person Re-Identification}.
\newblock {\em arXiv-Preprint}, 2017.

\bibitem{mipal}
Jihye Hwang, Jieun Lee, Sungheon Park, and Nojun Kwak.
\newblock {Pose estimator and tracker using temporal flow maps for limbs}.
\newblock In {\em IJCNN}, 2019.

\bibitem{10.5555/3045118.3045167}
Sergey Ioffe and Christian Szegedy.
\newblock {Batch Normalization: Accelerating Deep Network Training by Reducing
  Internal Covariate Shift}.
\newblock In {\em ICML}, 2015.

\bibitem{jin2019}
Sheng Jin, Wentao Liu, Wanli Ouyang, and Chen Qian.
\newblock {Multi-person Articulated Tracking with Spatial and Temporal
  Embeddings}.
\newblock {\em CVPR}, 2019.

\bibitem{DIP}
Dengjie Li, Siyu Chen, Yujie Zhong, Fan Liang, and Lin Ma.
\newblock {DiP: Learning Discriminative Implicit Parts for Person
  Re-Identification}.
\newblock In {\em arXiv-Preprint}, 2022.

\bibitem{li2019}
Jianing Li, Shiliang Zhang, and Tiejun Huang.
\newblock {Multi-Scale 3D Convolution Network for Video Based Person
  Re-Identification}.
\newblock In {\em AAAI}, 2019.

\bibitem{fpn}
Tsung-Yi Lin, Piotr Doll\'{a}r, Ross~B. Girshick, Kaiming He, Bharath
  Hariharan, and Serge~J. Belongie.
\newblock {Feature Pyramid Networks for Object Detection.}
\newblock In {\em CVPR}, 2017.

\bibitem{mscoco}
Tsung-Yi Lin, Michael Maire, Serge Belongie, James Hays, Pietro Perona, Deva
  Ramanan, Piotr Doll{\'a}r, and C.~Lawrence Zitnick.
\newblock {Microsoft COCO: Common Objects in Context}.
\newblock In {\em ECCV}, 2014.

\bibitem{adamw}
Ilya Loshchilov and Frank Hutter.
\newblock {Decoupled Weight Decay Regularization}.
\newblock In {\em ICCV}, 2019.

\bibitem{luiten2020IJCV}
Jonathon Luiten, Aljosa Osep, Patrick Dendorfer, Philip Torr, Andreas Geiger,
  Laura Leal-Taix{\'e}, and Bastian Leibe.
\newblock {HOTA: A Higher Order Metric for Evaluating Multi-Object Tracking}.
\newblock {\em IJCV}, 2020.

\bibitem{bagoftricks}
Hao Luo, Youzhi Gu, Xingyu Liao, Shenqi Lai, and Wei Jiang.
\newblock {Bag of Tricks and a Strong Baseline for Deep Person
  Re-Identification.}
\newblock In {\em CVPRW}, 2019.

\bibitem{spade}
Taesung Park, Ming{-}Yu Liu, Ting{-}Chun Wang, and Jun{-}Yan Zhu.
\newblock {Semantic Image Synthesis With Spatially-Adaptive Normalization}.
\newblock In {\em CVPR}, 2019.

\bibitem{DBLP:journals/corr/abs-1804-03864}
Lei Qi, Jing Huo, Lei Wang, Yinghuan Shi, and Yang Gao.
\newblock {MaskReID: A Mask Based Deep Ranking Neural Network for Person
  Re-identification}.
\newblock {\em ICME}, 2018.

\bibitem{STAF}
Yaadhav Raaj, Haroon Idrees, Gines Hidalgo, and Yaser Sheikh.
\newblock {Efficient Online Multi-Person 2D Pose Tracking with Recurrent
  Spatio-Temporal Affinity Fields}.
\newblock {\em CVPR}, 2019.

\bibitem{rafi2020selfsupervised}
Umer Rafi, Andreas Doering, Bastian Leibe, and Juergen Gall.
\newblock {Self-supervised Keypoint Correspondences for Multi-Person Pose
  Estimation and Tracking in Videos}.
\newblock In {\em ECCV}, 2020.

\bibitem{fasterRCNN}
Shaoqing Ren, Kaiming He, Ross~B. Girshick, and Jian Sun.
\newblock {Faster R-CNN: Towards Real-Time Object Detection with Region
  Proposal Networks.}
\newblock In {\em NeurIPS}, 2015.

\bibitem{sarfraz18}
M.~Saquib Sarfraz, Arne Schumann, Andreas Eberle, and Rainer Stiefelhagen.
\newblock {A Pose-Sensitive Embedding for Person Re-Identification with
  Expanded Cross Neighborhood Re-Ranking}.
\newblock {\em CVPR}, 2018.

\bibitem{Singh2018SNIPEREM}
Bharat Singh, Mahyar Najibi, and Larry~S. Davis.
\newblock {SNIPER: Efficient Multi-Scale Training}.
\newblock {\em NeurIPS}, 2018.

\bibitem{15_keypoints}
Michael Snower, Asim Kadav, Farley Lai, and Hans~Peter Graf.
\newblock {15 Keypoints Is All You Need}.
\newblock {\em CVPR}, 2020.

\bibitem{sun2019}
Yifan Sun, Qin Xu, Yali Li, Chi Zhang, Yikang Li, Shengjin Wang, and Jian Sun.
\newblock {Perceive Where to Focus: Learning Visibility-aware Part-level
  Features for Partial Person Re-identification}.
\newblock {\em CVPR}, 2019.

\bibitem{pcb}
Yifan Sun, Liang Zheng, Yi Yang, Qi Tian, and Shengjin Wang.
\newblock {Beyond Part Models: Person Retrieval with Refined Part Pooling (and
  A Strong Convolutional Baseline)}.
\newblock In {\em ECCV}, 2018.

\bibitem{7780677}
Christian Szegedy, Vincent Vanhoucke, Sergey Ioffe, Jon Shlens, and Zbigniew
  Wojna.
\newblock {Rethinking the Inception Architecture for Computer Vision}.
\newblock In {\em CVPR}, 2016.

\bibitem{transformer}
Ashish Vaswani, Noam Shazeer, Niki Parmar, Jakob Uszkoreit, Llion Jones,
  Aidan~N Gomez, {\L}ukasz Kaiser, and Illia Polosukhin.
\newblock {Attention is All you Need}.
\newblock In {\em NeurIPS}, 2017.

\bibitem{combining_detection_and_tracking}
Manchen Wang, Joseph Tighe, and Davide Modolo.
\newblock {Combining detection and tracking for human pose estimation in
  videos}.
\newblock {\em CVPR}, 2020.

\bibitem{realtimemot}
Zhongdao Wang, Liang Zheng, Yixuan Liu, and Shengjin Wang.
\newblock {Towards Real-Time Multi-Object Tracking}.
\newblock {\em ECCV}, 2020.

\bibitem{Wei2017GLADGD}
Longhui Wei, Shiliang Zhang, Hantao Yao, Wen Gao, and Qi Tian.
\newblock {GLAD: Global-Local-Alignment Descriptor for Pedestrian Retrieval}.
\newblock {\em ICM}, 2017.

\bibitem{Wieczorek2021OnTU}
Mikolaj Wieczorek, Barbara Rychalska, and Jacek Dabrowski.
\newblock On the unreasonable effectiveness of centroids in image retrieval.
\newblock In {\em ICNIP}, 2021.

\bibitem{simple_baseline}
Bin Xiao, Haiping Wu, and Yichen Wei.
\newblock {Simple Baselines for Human Pose Estimation and Tracking}.
\newblock {\em ECCV}, 2018.

\bibitem{learning_dynamics_via_graph}
Yiding Yang, Zhou Ren, Haoxiang Li, Chunluan Zhou, Xinchao Wang, and Gang Hua.
\newblock {Learning Dynamics via Graph Neural Networks for Human Pose
  Estimation and Tracking}.
\newblock {\em CVPR}, 2021.

\bibitem{10.1007/978-3-030-11012-3_19}
Dongdong Yu, Kai Su, Jia Sun, and Changhu Wang.
\newblock {Multi-person Pose Estimation for Pose Tracking with Enhanced
  Cascaded Pyramid Network}.
\newblock In {\em ECCVW}, 2018.

\bibitem{offset_guided_networks}
Rui Zhang, Zheng Zhu, Peng Li, Rui Wu, Chaoxu Guo, Guan Huang, and Hailun Xia.
\newblock {Exploiting Offset-guided Network for Pose Estimation and Tracking}.
\newblock In {\em {CVPRW}}, 2019.

\bibitem{zhang2021fairmot}
Yifu Zhang, Chunyu Wang, Xinggang Wang, Wenjun Zeng, and Wenyu Liu.
\newblock {FairMOT: On the Fairness of Detection and Re-Identification in
  Multiple Object Tracking}.
\newblock {\em ICCV}, 2021.

\bibitem{8693885}
Liang Zheng, Yujia Huang, Huchuan Lu, and Yi Yang.
\newblock {Pose-Invariant Embedding for Deep Person Re-Identification}.
\newblock {\em IEEE Transactions on Image Processing}, 2019.

\bibitem{zhou2020temporal}
Chunluan Zhou, Zhou Ren, and Gang Hua.
\newblock {Temporal keypoint matching and refinement network for pose
  estimation and tracking}.
\newblock In {\em ECCV}, 2020.

\bibitem{zhu2018deformable}
Xizhou Zhu, Han Hu, Stephen Lin, and Jifeng Dai.
\newblock {Deformable ConvNets v2: More Deformable, Better Results}.
\newblock {\em arXiv preprint arXiv:1811.11168}, 2018.

\end{thebibliography}
}

\end{document}